\documentclass[sigconf]{acmart}
\usepackage[table,HTML]{xcolor}
\usepackage{algorithm}
\usepackage{algorithmic}
\usepackage{listings}
\usepackage{xspace}
\usepackage{pbalance}
\usepackage{booktabs} 
\usepackage{multirow}
\usepackage{makecell}
\usepackage{ltablex} 

\usepackage{pifont} % 用来显示特殊符号
\usepackage[skins, breakable]{tcolorbox}
\usepackage{textcomp}

\AtBeginDocument{%
  }

\setcopyright{acmlicensed}
\copyrightyear{2018}
\acmYear{2018}
\acmDOI{XXXXXXX.XXXXXXX}
\acmConference[Conference acronym 'XX]{Make sure to enter the correct
  conference title from your rights confirmation email}{June 03--05,
  2018}{Woodstock, NY}
\acmISBN{978-1-4503-XXXX-X/2018/06}

\newcommand{\ourmethod}{\textsc{ViSAGE}\xspace}

\begin{document}

%%
%% The "title" command has an optional parameter,
%% allowing the author to define a "short title" to be used in page headers.
\title{\ourmethod: Constructing Self-Correcting Memories for Long-Form Video Understanding}
%%
%% The "author" command and its associated commands are used to define
%% the authors and their affiliations.
%% Of note is the shared affiliation of the first two authors, and the
%% "authornote" and "authornotemark" commands
%% used to denote shared contribution to the research.
\author{Xinkui Zhao}
\email{zhaoxinkui@zju.edu.cn}

\author{Enbo Chen}
\email{cebo@zju.edu.cn}

\author{Yifan Zhang}
\email{12451018@zju.edu.cn}

\author{Chang Liu}
\email{chang.liu@zju.edu.cn}

\author{Guanjie Cheng}
\email{chengguanjie@zju.edu.cn}

\author{Naibo Wang}
\email{wangnaibo@zju.edu.cn}

\affiliation{%
  \institution{Zhejiang University}
  \city{Hangzhou}
  \country{China}
}

\author{Yueshen Xu}
\email{ysxu@xidian.edu.cn}
\affiliation{%
  \institution{Xidian University}
  \city{Xi'an}
  \country{China}
}

%% 提示：加上下面这行可以修复页眉显示的错误（目前显示的是 Trovato et al.）
\renewcommand{\shortauthors}{Zhao et al.}

%%
%% The abstract is a short summary of the work to be presented in the
%% article.
\begin{abstract}
Multimodal agents operating in long-horizon environments must build and continually update multimedia memories to support entity-consistent, temporally grounded reasoning. However, existing agentic memory approaches often discard fine-grained identity cues under aggressive compression and segment-wise processing. They also rely heavily on vector-similarity retrieval, which can surface semantically related yet identity-mismatched evidence, leading to entity confusion, error propagation, and hallucinated answers. 
We propose \ourmethod, a multimodal agentic memory framework that constructs self-correcting, entity-centric memories. Specifically, \ourmethod anchors entity identity via cross-modal binding over long temporal ranges. It then applies bidirectional memory refinement to propagate delayed identity evidence, retroactively unifying historical records and improving future reasoning. We also introduce multi-agent cross-verification to assess retrieved evidence under an identity--evidence alignment constraint, enabling abstention instead of unsupported answers when evidence is missing. Extensive results demonstrate that \ourmethod consistently outperforms the strongest baseline, achieving 5.9\% higher accuracy.
\end{abstract}

%%
%% The code below is generated by the tool at http://dl.acm.org/ccs.cfm.
%% Please copy and paste the code instead of the example below.
%%
\begin{CCSXML}
<ccs2012>
   <concept>
       <concept_id>10010147.10010178.10010187</concept_id>
       <concept_desc>Computing methodologies~Knowledge representation and reasoning</concept_desc>
       <concept_significance>500</concept_significance>
       </concept>
 </ccs2012>
\end{CCSXML}

\ccsdesc[500]{Computing methodologies~Knowledge representation and reasoning}

%%
%% Keywords. The author(s) should pick words that accurately describe
%% the work being presented. Separate the keywords with commas.
\keywords{Multimodal Memory, Long-Form Video Understanding, Vision-Language Models, Self-Correction}
%% A "teaser" image appears between the author and affiliation
%% information and the body of the document, and typically spans the
%% page.

% \received{20 February 2007}
% \received[revised]{12 March 2009}
% \received[accepted]{5 June 2009}

%%
%% This command processes the author and affiliation and title
%% information and builds the first part of the formatted document.
\maketitle

\section{Introduction}
\begin{figure}
  \includegraphics[width=\linewidth]{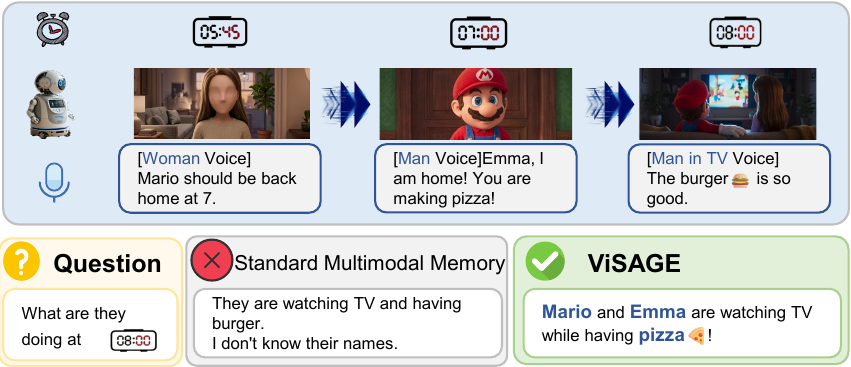}
  \caption{\textbf{Superior long-horizon reasoning with \ourmethod.} In complex long videos, standard memory struggles with delayed identities and audio-visual misalignment. \ourmethod effectively resolves these ambiguities. By establishing consistent entity tracking across the timeline, it bypasses misleading noise and delivers precise, hallucination-free answers to complex temporal queries.}
  \Description{\ourmethod processes raw video by extracting and aligning fragmented multimodal cues—using A(·) for audio analysis, F(·) for face detection, and L(·) for LLM-based appearance (looks) analysis —into chronological timelines. Through bidirectional memory refinement, it resolves anonymous entities into a unified identity (e.g., Alice), constructing a dual-track memory of sequential Incident Logs and profile-updating Object Cards. During query resolution, Epistemic Adjudication employs persona-driven multi-agent cross-verification. Agents retrieve evidence from the memory layers to ensure responses are factually grounded and identity-consistent, enabling verified refusals when an agent lacks relevant context.}
  \label{fig:teaser}
\end{figure}

Driven by the rapid progress of multimodal large language models (MLLMs)~\citep{gemini3_2025,lin2024vila,li2025videochat}, modern agents are now empowered to perceive and communicate through complex multimedia streams, seamlessly integrating vision, audio, and language~\citep{fan2024videoagent,fung2025embodied}.
However, deploying such agents in long-horizon settings requires memory that persists beyond a single context window. 
Therefore, prior work typically uses retrieval-augmented memory to extend MLLMs beyond finite context windows~\citep{park2023generative,zhong2024memorybank,chhikara2025mem0,liu2024agentlite,hu2025hiagent}. 
% In this paradigm, the agent writes observations and intermediate states into an external memory, and uses the current dialogue state or task context as a query to retrieve relevant entries as additional evidence for generation and decision making. 
In this paradigm, the agent writes observations to external memory and retrieves relevant entries to guide generation and decisions.
% Many systems implement memory as a vector store and rely on embedding-based retrieval for scalability
Many systems use vector-store retrieval~\citep{zhong2024memorybank,chhikara2025mem0,fan2024videoagent,fan2025embodied}, while hierarchical designs organize memories across multiple time scales for long-horizon planning~\citep{liu2024agentlite,hu2025hiagent}. Optimus-1~\citep{li2024optimus} further explores hybrid multimodal memory, combining a hierarchical directed knowledge graph with an abstracted experience pool to encode world knowledge and past multimodal experience. More recently, M3-Agent structures multimodal memory as an entity-centric graph to better support long-horizon reasoning over entities~\citep{long2025seeing}.

Although these efforts have shown promising performance in long-horizon multimodal agentic memory tasks, we identify several limitations in current paradigms:
\begin{itemize}
    \item \textbf{Spatio-Temporal Detail Loss.} Existing systems face a trade-off between span and detail. To ingest extensive multimodal streams, they often rely on aggressive down-sampling or compression, which erases critical micro-dynamics and thus removes fine-grained identity cues needed for reliable entity disambiguation.

    \item \textbf{Segmentation Dilemma.} Many long-horizon pipelines process inputs in isolated chunks. 
    % However, key identity information is often revealed only later in the stream, and effective understanding requires linking that later evidence back to earlier events.
    Identity evidence often arrives late and must be linked back to earlier events.
    % Chunk-wise processing makes such cross-segment association difficult, so identity cues discovered downstream cannot reliably revise or consolidate earlier memories.
    Chunk-wise processing blocks cross-segment revision, so late cues cannot fix earlier memories.

    \item \textbf{Similarity--Veracity Gap.} Retrieval-based memory conflates semantic relevance with factual correctness. Vector-similarity retrieval can surface contextually related yet identity-mismatched or incorrect evidence, without verification, agents are prone to answering based on noisy retrieval.
\end{itemize}

\begin{figure*}[htbp] 
  \centering 
  \includegraphics[width=0.9\textwidth]{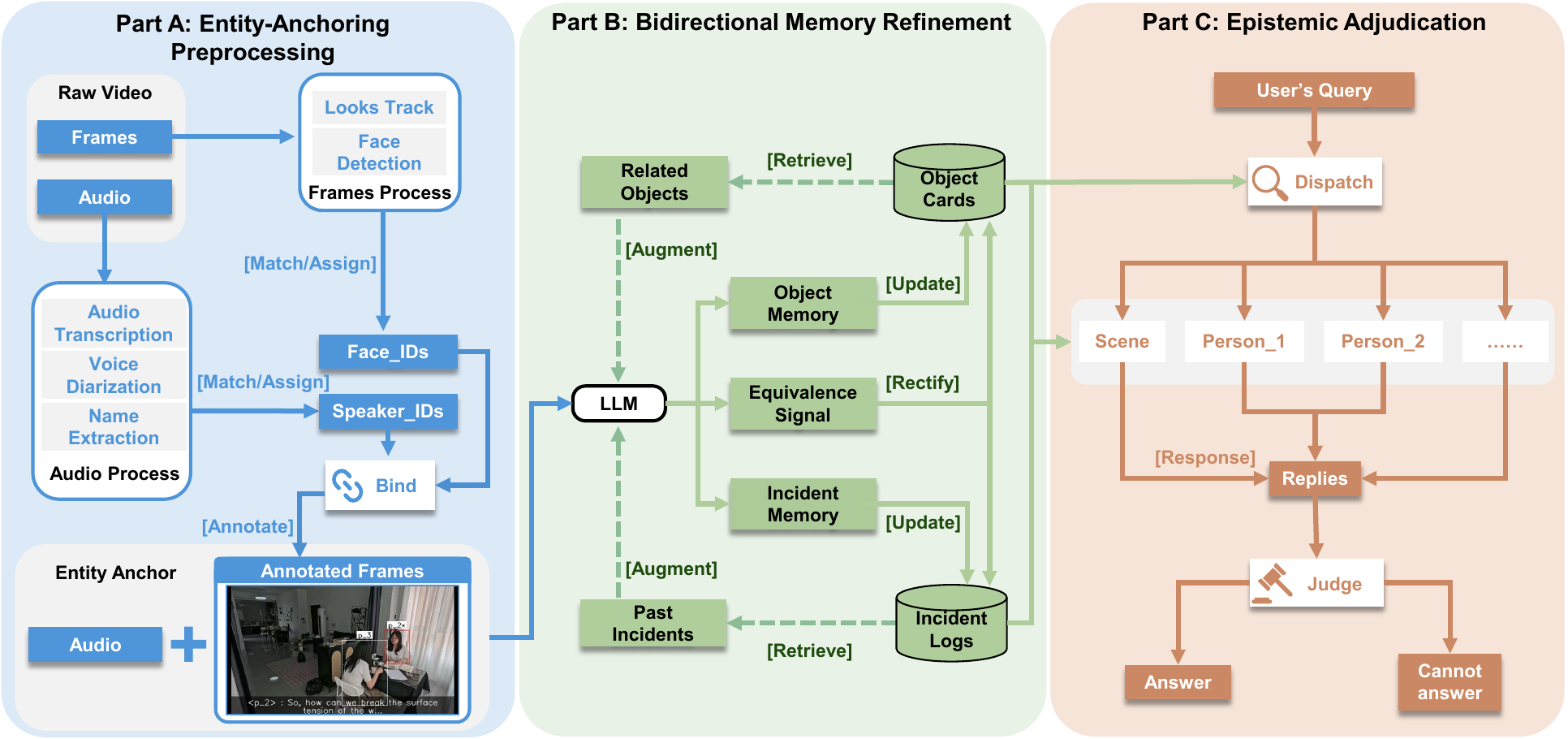} 
  \caption{Architecture of \ourmethod. The framework proceeds in three stages: (A) Entity-Anchoring Preprocessing binds visual and audio signals to annotated frames; (B) Bidirectional Context Refinement maintains self-correcting memory (Incident Logs and Object Cards) via retrieval and rectification; and (C) Epistemic Adjudication employs multi-agent cross-verification to ensure reliable responses.}
  \Description{Architecture of \ourmethod. The framework proceeds in three stages: (A) Entity-Anchoring Preprocessing binds visual and audio signals to annotated frames; (B) Bidirectional Context Refinement maintains self-correcting memory (Incident Logs and Object Cards) via retrieval and rectification; and (C) Epistemic Adjudication employs multi-agent cross-verification to ensure reliable responses.}
\label{fig:method}
\end{figure*}

To address these challenges, we introduce \ourmethod, a \textbf{Vi}sual \textbf{S}elf-correcting \textbf{AGE}ntic memory framework that integrates fragmented multimedia signals to construct \emph{entity-centric, self-correcting} memories, transcending conventional linear, chunk-wise processing.
\ourmethod maintains a dual-structured memory: \emph{Incident Logs} record sequential narrative events, while \emph{Object Cards} maintain evolving entity profiles that accumulate and revise identity attributes.
To align identity cues across spatial degradation and temporal sparsity, we perform \textbf{Cross-Modal Entity Binding} by binding long-range visual tracklets with complementary multi-modal signals. 
To overcome isolated segmentation, \ourmethod further introduces \textbf{Bidirectional Memory Refinement} to propagate delayed identity evidence, retroactively unifying historical records and stabilizing future reasoning. 
Finally, we propose \textbf{Multi-Agent Cross-Verification} to adjudicate retrieved evidence under an identity--evidence alignment constraint, enabling verified refusals instead of hallucinations when evidence is missing.

We evaluate \ourmethod through extensive experiments on long-horizon multimodal benchmarks, including M3-Bench-robot, M3-Bench-web~\citep{long2025seeing}, and Video-MME-long~\citep{fu2025video}. Experimental results show that \ourmethod consistently outperforms strong baselines across diverse reasoning categories by 5.9\%, demonstrating robust gains on identity-critical queries. Moreover, our results verify that Multi-Agent Cross-Verification improves reliability by enforcing identity-evidence alignment, enabling verified refusals instead of hallucinated answers when supporting evidence is missing or noisy.
Our contributions are summarized as follows:

\begin{itemize}
    \item \textbf{Framework Design:} We introduce \ourmethod, a visual self-correcting agentic memory framework that enables \emph{retroactive, entity-centric} updates beyond chunk-wise processing. \ourmethod separates memory into sequential \emph{Incident Logs} and structured \emph{Object Cards}, allowing delayed identity evidence to be aligned with historical events for long-horizon entity consistency.

    \item \textbf{Algorithmic Novelty:} We propose \textbf{Bidirectional Memory Refinement} to resolve the Segmentation Dilemma. This mechanism transcends sequential processing via a feedback loop that retroactively unifies fragmented narratives using delayed identity evidence, ensuring long-term entity consistency.
    
    \item \textbf{Epistemic Safety:} To bridge the Similarity-Veracity Gap, we introduce \textbf{Multi-Agent Cross-Verification}. This framework enforces identity-constrained reasoning to mitigate hallucinations, prioritizing Verified Epistemic Refusals over plausible fabrications in high-stakes embodied scenarios.
    
    \item \textbf{Empirical Impact:} \ourmethod sets new state-of-the-art on authoritative benchmarks. Our framework consistently outperforms baselines, achieving accuracy of 45.5\% on M3-Bench-robot, 58.4\% on M3-Bench-web, and 79.1\% on Video-MME-long, validating its robustness in long-form video understanding.
\end{itemize}

\section{Related Work}

\subsection{Long-Form Video Understanding and the Context Bottleneck}
Recent advances in Multimodal Large Language Models (MLLMs), such as GPT-4o~\citep{hurst2024gpt}, Gemini-3-Pro~\citep{gemini3_2025}, and open-source Qwen3-VL~\citep{qwen3vl_2025}, have revolutionized visual perception. While methods like Video-LLaVA~\citep{lin2024video} and VILA~\citep{lin2024vila} excel in processing short clips, extending their capabilities to hour-long videos remains a significant challenge due to strict context window limits. Early solutions to this bottleneck typically followed a Socratic, text-based paradigm. Inspired by Socratic Models~\citep{zeng2022socratic}, these approaches decompose complex video reasoning into manageable, text-based tasks. Methods like Video Recap~\citep{islam2024video} and AutoAD~\citep{han2023autoad} generate hierarchical captions or dense summaries, effectively transforming the video into a format suitable for standard textual Retrieval-Augmented Generation (RAG) systems~\citep{lewis2020retrieval}. However, recent demanding benchmarks like EgoSchema~\citep{mangalam2023egoschema} and LongVideoBench~\citep{wu2024longvideobench} highlight a critical flaw: while computationally efficient, these translation-based methods suffer from severe information loss. By discarding fine-grained, continuous visual cues that are essential for deep temporal reasoning, text-only representations fail to capture the full nuance of long-horizon video data.

\subsection{Visual Context Expansion via Token Compression}
To address the severe information loss inherent in text-based translation, a subsequent line of work has focused on expanding model capacity to retain extensive visual features directly. To mitigate the overwhelming spatial-temporal redundancy of video data, the primary approach relies on token compression methods. To tackle severe memory consumption, models such as LongVILA~\citep{chen2024longvila} and LongVU~\citep{shen2024longvu} utilize spatiotemporal pooling mechanisms. LongVILA introduces system-level multi-modal sequence parallelism, whereas LongVU employs a spatiotemporal adaptive compression mechanism guided by cross-modal queries to intelligently discard task-irrelevant background frames. Furthermore, to address the inflexibility of fixed compression ratios, VidCompress~\citep{lan2024vidcompress} and Video-XL~\citep{liu2025video} employ adaptive selection. VidCompress dynamically determines the compression rate based on visual complexity, and Video-XL utilizes visual summarization tokens to seamlessly fit hour-scale videos into standard context windows.

\subsection{Memory-Centric Architectures and AI Agents}
Rather than compressing inputs to fit fixed context windows, an alternative paradigm introduces explicit external banks or buffers. These memory-centric models adopt a streaming or chunk-based processing approach to bypass linear memory growth. For instance, MovieChat~\citep{song2024moviechat} proposes a sparse memory mechanism for long-term storage. To address the lack of long-range spatiotemporal correlations and prevent catastrophic forgetting, MA-LMM~\citep{he2024ma} and LifelongMemory~\citep{wang2023lifelongmemory} utilize online memory banks and episodic buffers for active retrieval. At the architectural level, MeMViT~\citep{wu2022memvit} caches previous network activations, while Flash-VStream~\citep{zhang2024flash} optimizes for real-time updates with fast memory eviction for instantaneous video stream understanding.

Parallel to these architectural improvements in MLLMs, the evolution of AI Agents capable of autonomous, long-horizon planning has heavily relied on robust memory systems. Foundational works like Generative Agents~\citep{park2023generative} established the importance of retrieving past experiences. This has been formalized into scalable vector memory layers like Mem0~\citep{chhikara2025mem0} and MemoryBank~\citep{zhong2024memorybank}. Furthermore, hierarchical and operating-system-level memory structures, explored in HiAgent~\citep{hu2025hiagent}, AgentLite~\citep{liu2024agentlite}, and AIOS~\citep{mei2024aios}, enable complex, multi-step planning tasks.

Crucially, extending these agentic systems to the multimedia domain requires sophisticated memory integration. VideoAgent~\citep{fan2024videoagent} and its embodied variant Embodied VideoAgent~\citep{fan2025embodied} utilize external tools for iterative video retrieval, while Jarvis-1~\citep{wang2024jarvis} seamlessly integrates memory with multimodal planning. Approaches like M3-Agent~\citep{long2025seeing} even structure memory into an entity-centric graph. However, as these memory-augmented systems scale to handle increasingly dense multimodal inputs, a critical emerging challenge lies in constructing self-correcting memories that can dynamically refine, update, and disambiguate retrieved visual contexts over time, thereby preventing cascading errors during long-form video understanding tasks.

\section{Method}
As shown in Fig.~\ref{fig:method}, long-form video understanding requires integrating visual, auditory, and textual cues to maintain entity-consistent reasoning over time. To address identity fragmentation and unsupported inference caused by segment-wise processing, \ourmethod adopts a three-stage multimodal memory framework.

\subsection{Memory Bank}
\label{sec:memory}

Inspired by Tulving's multiple memory systems theory~\cite{tulving2002episodic}, which distinguishes time- and context-bound experience from decontextualized knowledge about entities and emphasizes a mechanistic link ``from mind to brain'', we design \ourmethod's memory bank as an incident--entity pair. Specifically, \emph{Incident Logs} store an append-only record of \emph{what happened}, including timestamps, actions, and dialogue. This event content is immutable, but the \emph{who} field remains refinable. In parallel, \emph{Object Cards} maintain continuously updated entity knowledge. This separation enables retroactive identity unification without corrupting the historical event record.

\begin{figure*}[!htb] 
  \centering 
  \includegraphics[width=\textwidth]{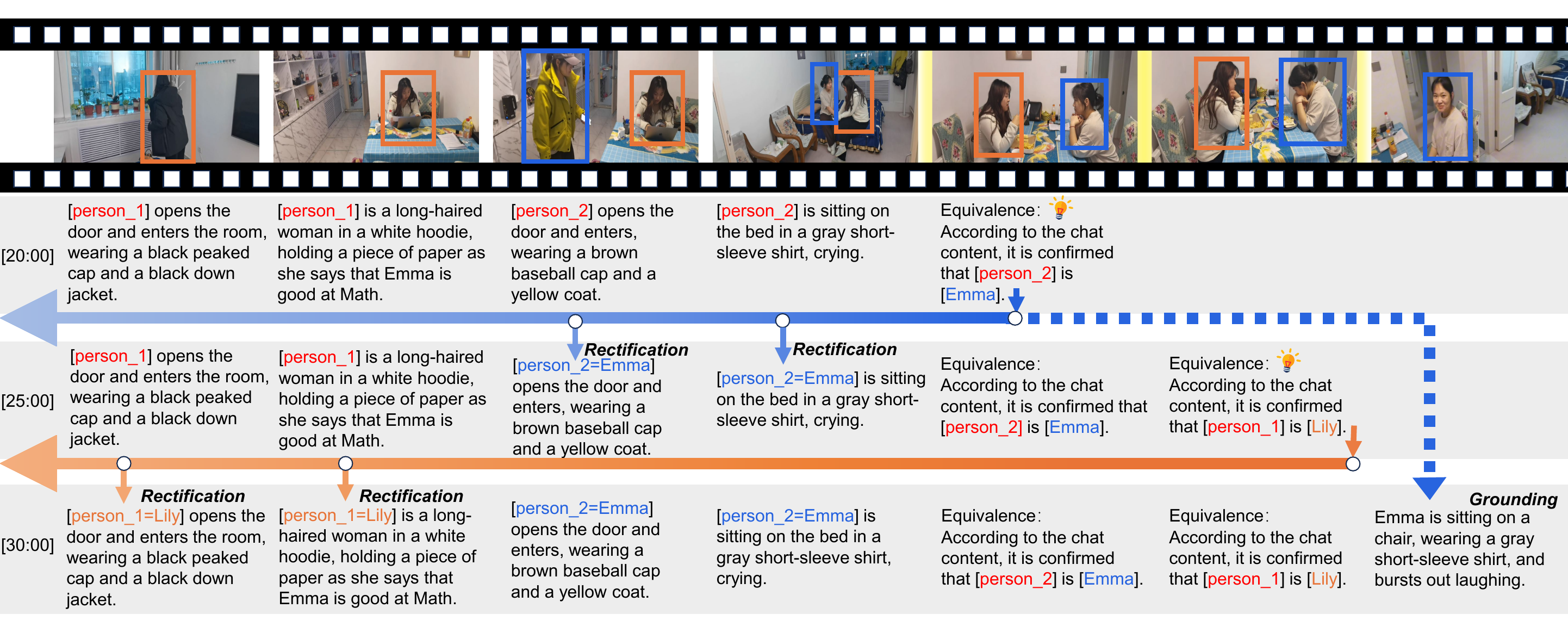} 
  \caption{Visualizing Bidirectional Memory Refinement. Late-appearing Identity Triggers activate a dual loop: Backward Rectification retroactively fills historical uncertainties, while Forward Grounding propagates these resolved identities to future timestamps. This ensures actions are anchored to the correct character profile.}
  \Description{Visualizing Bidirectional Memory Refinement. Late-appearing Identity Triggers activate a dual loop: Backward Rectification retroactively fills historical uncertainties, while Forward Grounding propagates these resolved identities to future timestamps. This ensures actions are anchored to the correct character profile.}
\label{fig:bid}
\end{figure*}

\subsection{Entity-Anchored Preprocessing}
\label{sec:entity}

\paragraph{Countering Spatial Degradation via Sequence-Level Entity Anchoring.}
Long-form videos are commonly processed under sparse frame sampling and compressed visual inputs to control computational cost. However, these settings often weaken fine-grained appearance cues and make identity recognition within a local segment unreliable, especially under occlusion, pose variation, and motion blur. Segment-wise processing therefore tends to produce identity switches and noisy event--entity associations during memory construction. To strike an optimal balance between performance and computational cost, our guiding principle is to compensate for this sampling-induced information loss by leveraging computationally cheaper modal cues without increasing the visual frame sampling rate. Consequently, \ourmethod maintains memory through different cost hierarchies: binding cross-modal cues during initial sensory encoding, performing forward grounding during active processing, and executing backward rectification upon memory retrieval.

The representative agent-based approach~\citep{long2025seeing} adopts a partition-and-process paradigm that slices long videos into independent segments. This confines identity recognition to each local window, making it brittle under occlusion, pose changes, or compression artifacts. As a result, long videos often attenuate fine-grained spatio-temporal cues, leading to identity switches and noisy event--entity associations during memory construction. Such errors propagate to both \emph{Incident Logs} and \emph{Object Cards}, degrading retrieval and consolidation. We therefore perform entity-anchored preprocessing to stabilize identity grounding at the sequence level.

To this end, we introduce \emph{Sequence-Level Entity Anchoring}, which replaces unstable per-frame identity assignment with track-level identity grounding. We first run multi-object tracking over the full video to obtain long-range person trajectories, and collect all face detections associated with each trajectory. For every trajectory, we select the highest-quality face instance as a \emph{prototype}, which serves as the canonical identity reference for that track. The remaining face observations, including partial or degraded ones, are then matched to the prototype and assigned the same entity label whenever the match is valid. This procedure converts a temporally fragmented set of face observations into a stable identity cluster. Each cluster is mapped to a persistent entity record in memory: the prototype initializes the corresponding \emph{Object Card}, and the assigned entity ID is used when writing event-related evidence into \emph{Incident Logs}. As a result, transient or low-quality visual cues can still be linked to the correct entity through the prototype, improving both event--entity association and attribute consolidation across the video.

\paragraph{Bridging Temporal Sparsity via Speaker Binding.}
Parallel to visual anchoring, we leverage audio cues to improve \emph{who-did-what} attribution, where knowing \emph{who is speaking} is decisive. A practical challenge is that standard MLLM pipelines operate on temporally sparsified video inputs, which discard the rapid lip dynamics needed for reliable speaker grounding. Meanwhile, relying solely on speaker embeddings~\cite{long2025seeing} can be brittle in crowded scenes due to acoustic similarity and intra-speaker variability. Motivated by the McGurk effect~\citep{mcgurk1976hearing}, which illustrates how vision can disambiguate ambiguous audio, we introduce \emph{Dual-Criteria Validation} for voice--face binding. Concretely, we run Active Speaker Detection (ASD) on the high-FPS stream to obtain per-face articulation/synchrony evidence, and combine it with acoustic matching from voice embeddings. We bind a \texttt{Voice\_ID} to a \texttt{Face\_ID} whenever the audio is \emph{confidently identifiable} with high voice-similarity or the video is \emph{physically corroborative} with ASD-confirmed audio--visual synchrony within the same time window; otherwise, the speaker remains unassigned. 

\subsection{Bidirectional Memory Refinement}
\label{sec:bi}

To address the limitation of segment-wise processing, where identity evidence may emerge later than the corresponding visual observation, \ourmethod introduces Bidirectional Memory Refinement. Unlike conventional unidirectional memory updates that keep earlier event--entity bindings fixed, this module combines Forward Grounding for current entity assignment with Retroactive Rectification, which revises prior Incident Logs using newly acquired identity evidence, as illustrated in Fig.~\ref{fig:bid}.

\paragraph{Forward Grounding via Identity-Keyed Retrieval.}
To counteract local contextual drift in the current timeframe, we implement an anchor-based retrieval strategy.
Prior to memory generation, the system utilizes the stable \texttt{Face\_ID}, derived from the tracklet anchoring phase, to retrieve memory from two parallel knowledge streams: the immediately preceding events from Incident Logs (for narrative continuity) and the character's accumulated profile from Object Cards (for persona consistency).
By priming the MLLM with this dual-source memory, we transform the interpretation of the current clip from an isolated fragment into a continuous narrative extension, ensuring the ``Now'' is grounded in the ``Past''.

\paragraph{Backward Refinement via Logical Merging.}
Despite the robust physical continuity established in Section~\ref{sec:entity}, where the system correctly binds the audio information to the entity, the agent remains blind to the character's nominal identity.
The mapping between a face and a name typically relies on complex dialogue comprehension, specifically determining the addressee in a conversation.
To handle this, we treat every newly appearing name as a provisional semantic identity.
The LLM performs memory-aware inference: if it deduces that a specific visual subject is the target of a naming utterance, such as ``Hey, Mario!'', it generates a Refinement Signal ($ID_{sem} \equiv ID_{vis}$). Crucially, this signal triggers a Global Backward Update: the system traverses the historical Incident Logs $\mathcal{L}$, retroactively replacing all prior instances of the anonymous $ID_{vis}$ with the resolved identity. This mechanism ensures that if an anonymous subject recorded on Day 1 is identified as ``Mario'' on Day 2, their entire history is unified, repairing the structural fragmentation.

\paragraph{Comparison with Existing Agentic Memories.}
This retroactive rectification establishes a fundamentally different paradigm compared to recent agentic memory systems. For instance, A-Mem~\citep{xu2025mem} employs a semantic-driven memory correction that identifies similarities between new and old memories to establish linkages, primarily optimizing for multi-hop reasoning. In contrast, \ourmethod's refinement is strictly evidence-driven: it triggers updates exclusively upon discovering definitive identity evidence, explicitly modifying the historical content within both Object Cards and Incident Logs rather than merely linking them. This directly resolves identity confusion and hallucinations inherent in information-lossy long video understanding. Furthermore, while Optimus-1~\citep{li2024optimus} maintains an accumulative memory structure---storing static execution logs and retrieving past failures to help the agent avoid repeating mistakes in new tasks---\ourmethod is designed to actively reach back and correct the underlying database based on new evidence. This ensures the agent continually operates on a disambiguated, highly definitive factual record rather than a static accumulation of past states.

\begin{table*}[t]
    \centering
    % 【核心修改 1】使用合法的 \small 命令（8pt），彻底抛弃 \resizebox
    \small 
    \caption{
    Performance Comparison on M3-Bench and Video-MME-long. 
    We report results across different types in M3-Bench: 
    Multi-Evidence (ME), 
    Multi-Hop (MH), 
    Cross-Modal (CM), 
    Person Understanding (PU), 
    and General Knowledge (GK). 
    }
    \Description{Performance Comparison on M3-Bench and Video-MME-long. 
    We report results across different types in M3-Bench: 
    Multi-Evidence (ME), 
    Multi-Hop (MH), 
    Cross-Modal (CM), 
    Person Understanding (PU), 
    and General Knowledge (GK). }
    \renewcommand{\arraystretch}{1.2} % 保持行高舒适
    \setlength{\tabcolsep}{3pt}       % 缩紧列间距，确保 14 列能安全塞进页面
    
    % 【核心修改 2】直接使用标准的 tabular 环境，让表格自然排版、居中对齐
    \begin{tabular}{lccccccccccccc}
        \hline
        \multirow{2}{*}{\textbf{Method}} & 
        \multicolumn{6}{c}{\textbf{M3-Bench-robot}} & 
        \multicolumn{6}{c}{\textbf{M3-Bench-web}} & 
        \multirow{2}{*}{\textbf{\shortstack{Video-\\MME-long}}} \\
        
        \cline{2-7} \cline{8-13}
        
         & ME & MH & CM & PU & GK & ALL & ME & MH & CM & PU & GK & ALL & \\
        \hline
        
        \multicolumn{14}{c}{\textit{Video-Native Socratic method}} \\ 
        \hline
        Qwen2.5-Omni-7b & 2.1 & 1.4 & 1.5 & 1.5 & 2.1 & \cellcolor[HTML]{E6F3FF}2.0 & 8.9 & 8.8 & 13.7 & 10.8 & 14.1 & \cellcolor[HTML]{E6F3FF}11.3 & \cellcolor[HTML]{E6F3FF}42.2 \\
        Gemini-1.5-Pro & 6.5 & 7.5 & 8.0 & 9.7 & 7.6 & \cellcolor[HTML]{E6F3FF}8.0 & 18.0 & 17.9 & 23.8 & 23.1 & 28.7 & \cellcolor[HTML]{E6F3FF}23.2 & \cellcolor[HTML]{E6F3FF}38.1 \\
        \hline
        
        \multicolumn{14}{c}{\textit{Online Video Understanding Methods}} \\
        \hline
        MovieChat & 13.3 & 9.8 & 12.2 & 15.7 & 7.0 & \cellcolor[HTML]{E6F3FF}11.2 & 12.2 & 6.6 & 12.5 & 17.4 & 11.1 & \cellcolor[HTML]{E6F3FF}12.6 & \cellcolor[HTML]{E6F3FF}19.4 \\
        MA-LMM & 25.6 & 23.4 & 22.7 & 39.1 & 14.4 & \cellcolor[HTML]{E6F3FF}24.4 & 26.8 & 10.5 & 22.4 & 39.3 & 15.8 & \cellcolor[HTML]{E6F3FF}24.3 & \cellcolor[HTML]{E6F3FF}17.3 \\
        Flash-Vstream & 21.6 & 19.6 & 19.3 & 24.3 & 14.1 & \cellcolor[HTML]{E6F3FF}19.4 & 24.5 & 10.3 & 24.6 & 32.5 & 20.2 & \cellcolor[HTML]{E6F3FF}23.6 & \cellcolor[HTML]{E6F3FF}25.0 \\
        \hline
        
        \multicolumn{14}{c}{\textit{Discrete Socratic method}} \\ 
        \hline
        Qwen3-VL-8b & 39.0 & 29.7 & 39.3 & 52.9 & 18.8 & \cellcolor[HTML]{E6F3FF}36.1 & 38.8 & 22.0 & 40.0 & 40.5 & 45.1 & \cellcolor[HTML]{E6F3FF}36.9 & \cellcolor[HTML]{E6F3FF}60.1 \\
        Qwen3-VL-30b & 36.8 & 36.1 & 37.1 & 53.3 & 22.2 & \cellcolor[HTML]{E6F3FF}36.4 & 43.0 & 26.0 & 40.0 & 50.4 & 45.1 & \cellcolor[HTML]{E6F3FF}40.9 & \cellcolor[HTML]{E6F3FF}67.0 \\
        GPT-4o & 34.0 & 32.8 & 32.9 & 42.2 & 16.7 & \cellcolor[HTML]{E6F3FF}29.9 & - & - & - & - & - & \cellcolor[HTML]{E6F3FF}- & \cellcolor[HTML]{E6F3FF}65.3 \\
        Gemini-3-Pro & 42.7 & 38.3 & 38.0 & 55.1 & 23.3 & \cellcolor[HTML]{E6F3FF}39.6 & 51.9 & 40.3 & 62.9 & 61.0 & 59.0 & \cellcolor[HTML]{E6F3FF}53.8 & \cellcolor[HTML]{E6F3FF}74.2 \\
        \hline
        
        \multicolumn{14}{c}{\textit{Agent-based Method}} \\ 
        \hline
        M3-Agent & 32.8 & 29.4 & 31.2 & 43.3 & 19.1 & \cellcolor[HTML]{E6F3FF}30.7 & 45.9 & 28.4 & 44.3 & 59.3 & 53.9 & \cellcolor[HTML]{E6F3FF}48.9 & \cellcolor[HTML]{E6F3FF}61.8 \\
        \textbf{\ourmethod(Ours)} & \textbf{45.2} & \textbf{47.2} & \textbf{47.1} & \textbf{56.9} & \textbf{30.9} & \cellcolor[HTML]{E6F3FF}\textbf{45.5} & \textbf{62.3} & \textbf{47.8} & \textbf{71.4} & \textbf{63.3} & \textbf{62.2} & \cellcolor[HTML]{E6F3FF}\textbf{58.4} & \cellcolor[HTML]{E6F3FF}\textbf{79.1} \\
        \hline
    \end{tabular}

    \label{tab:main}
\end{table*}

\subsection{Epistemic Adjudication}
\label{sec:adjudication} % 建议加个 label

To bridge the Similarity--Veracity Gap, \ourmethod introduces Object Cards as a high-confidence knowledge anchor, supplementing the traditional retrieval from Incident Logs.

\paragraph{Noise Filtering via Identity Resolution Dispatcher.} 
To identify the entity most relevant to the query, the Dispatcher functions as a sanitization layer against ASR-induced noise. Automatic transcription frequently introduces phonetic inconsistencies, such as erroneously transcribing the target ``Mario'' as ``Malloy'', which causes naive vector retrieval to fail or return irrelevant distractors. Capitalizing on the fact that all valid character names are \emph{indexed} in the database, the Dispatcher employs LLM-based logical inference to perform Identity Resolution. It explicitly maps the noisy query entity to its correct counterpart within this registry, effectively correcting errors like ``Malloy'' to ``Mario''. It then activates only the relevant Character Agent, ensuring that retrieval is strictly targeted and robust to phonetic mismatches.

\paragraph{Evidence Extraction via Persona-Driven Agents.}
Crucially, \ourmethod employs a complementary retrieval strategy specialized by query type. The Scene Agent interrogates Incident Logs to answer factual, event-driven questions. It utilizes the objective chronological record to resolve dynamic narrative details, such as ``who stood up first'', which rely on precise temporal causality. Complementing this, the Character Agent leverages Object Cards to resolve static attribute inquiries. Without this solidified profile, answering identity-related questions like ``What is Lily's profession?'' would necessitate on-the-fly inference based on disjointed event fragments in the Incident Logs. This limitation is evident in accumulative memory graphs like M3-Agent~\citep{long2025seeing}. When asked ``What is Lily's favorite drink?'', such systems rely on semantic similarity to first retrieve anonymous nodes (e.g., ``character\_1 likes yogurt''), necessitating multiple, error-prone RAG rounds to subsequently deduce that ``character\_1'' is indeed Lily. Because \ourmethod's retroactive refinement (Section~\ref{sec:bi}) has already directly rewritten and disambiguated the underlying database, our persona-driven agents can fetch precise, identity-linked answers in a single step. This eliminates redundant retrieval rounds, significantly accelerating response speed and improving retrieval accuracy.

\paragraph{Epistemic Adjudication via Judge Agent.} 
In the final phase, the Judge Agent functions as the ultimate arbiter of veracity, addressing a critical flaw in existing works like M3-Agent~\citep{long2025seeing} that predominantly rely on pure semantic similarity for retrieval. Such approaches falsely assume that high semantic overlap equates to valid evidence. In practice, this susceptibility to semantic noise causes severe hallucinations; for example, retrieving hearsay where others repeatedly discuss ``Lily being good at math'' might mislead the agent into confidently hallucinating that her major is mathematics. Instead of simply aggregating outputs, \ourmethod utilizes identity-based memory filtering to discard irrelevant information, and the Judge Agent rigorously evaluates the actual contribution, sufficiency, and consistency of the evidence retrieved from both Incident Logs and Object Cards. Crucially, when valid evidence is absent in both sources, or when irreconcilable conflicts arise, the Judge triggers a Verified Epistemic Refusal. This strict rejection mechanism ensures safety by preventing the system from forcing answers to unanswerable queries, establishing a more trustworthy agentic framework.

\section{Experiments}
\label{sec:experiments}

\subsection{Experimental Setup}
\begin{table}[htbp] % 假设你外面包了 table 环境
    \centering
    \small % 建议加个 small，让表格更精致
    \caption{Dataset Statistics} % 别忘了加个 caption
    \begin{tabular}{lccc}
        \toprule % <--- 第一个 \hline 替换为 \toprule (封顶)
        \textbf{Dataset Split} & \textbf{robot} & \textbf{web} & \textbf{V-MME-long} \\
        \midrule % <--- 第二个 \hline 替换为 \midrule (全场唯一，隔离表头与数据)
        
        \textbf{\# Videos} & 100 & 920 & 300 \\
        % 这两行数据之间不需要加线，如果觉得太挤，可以加一句 \addlinespace
        \textbf{Avg. Length} & 34m & 27m & 40m \\
        \bottomrule % <--- (封底)
    \end{tabular}
    \label{tab:datasets}
\end{table}

\paragraph{Evaluation Datasets.} 
To comprehensively analyze \ourmethod, we utilize three authoritative benchmarks (summarized in Table~\ref{tab:datasets}). 
First, the robot split of M3-Bench~\citep{long2025seeing} focuses on general-purpose robots, featuring egocentric videos that test memory-guided reasoning such as inferring human personalities, interpersonal relationships, and object affordances. 
Complementing this, the web split of M3-Bench~\citep{long2025seeing} and Video-MME~\citep{fu2025video} provide content with high information density from online platforms. 
These videos cover diverse topics like documentaries and movies, challenging the agent to process complex narratives and open-world knowledge relevant to practical multimodal applications.

\paragraph{Baselines.} 
We benchmark \ourmethod against a suite of methods across three paradigms: 
(1) \textbf{Socratic Models}, encompassing Video-Native systems (Gemini-1.5-Pro~\citep{team2024gemini}, Qwen2.5-Omni~\citep{xu2025qwen2}) that process holistic streams, and Discrete variants (GPT-4o~\citep{hurst2024gpt}, Qwen3-VL~\citep{qwen3vl_2025}, Gemini-3-Pro~\citep{gemini3_2025}) utilizing frames sampled at 0.5 fps with ASR; 
(2) \textbf{Online Video Understanding Methods} representing memory compression strategies, including MovieChat~\citep{song2024moviechat}, MA-LMM~\citep{he2024ma}, and Flash-VStream~\citep{zhang2024flash}; 
and (3) \textbf{Agent-based Frameworks}, specifically the previous SOTA M3-Agent~\citep{long2025seeing}, which employs an entity-centric memory graph and serves as the primary structural benchmark.

\paragraph{Implementation Details.} 
We utilize Qwen3-Omni-Flash~\citep{xu2025qwen3omni} for speech transcription, followed by Gemini-3-Pro as the backbone MLLM for reasoning and memory synthesis. 
% See Appendix~\ref{app:appendix_impl} for more details.
%The complete set of prompt templates used in our experiments is provided in \textbf{Appendix~\ref{sec:appendix_prompts}}.

\begin{figure}[t]
    % \centering
    \includegraphics[width=0.9\columnwidth]{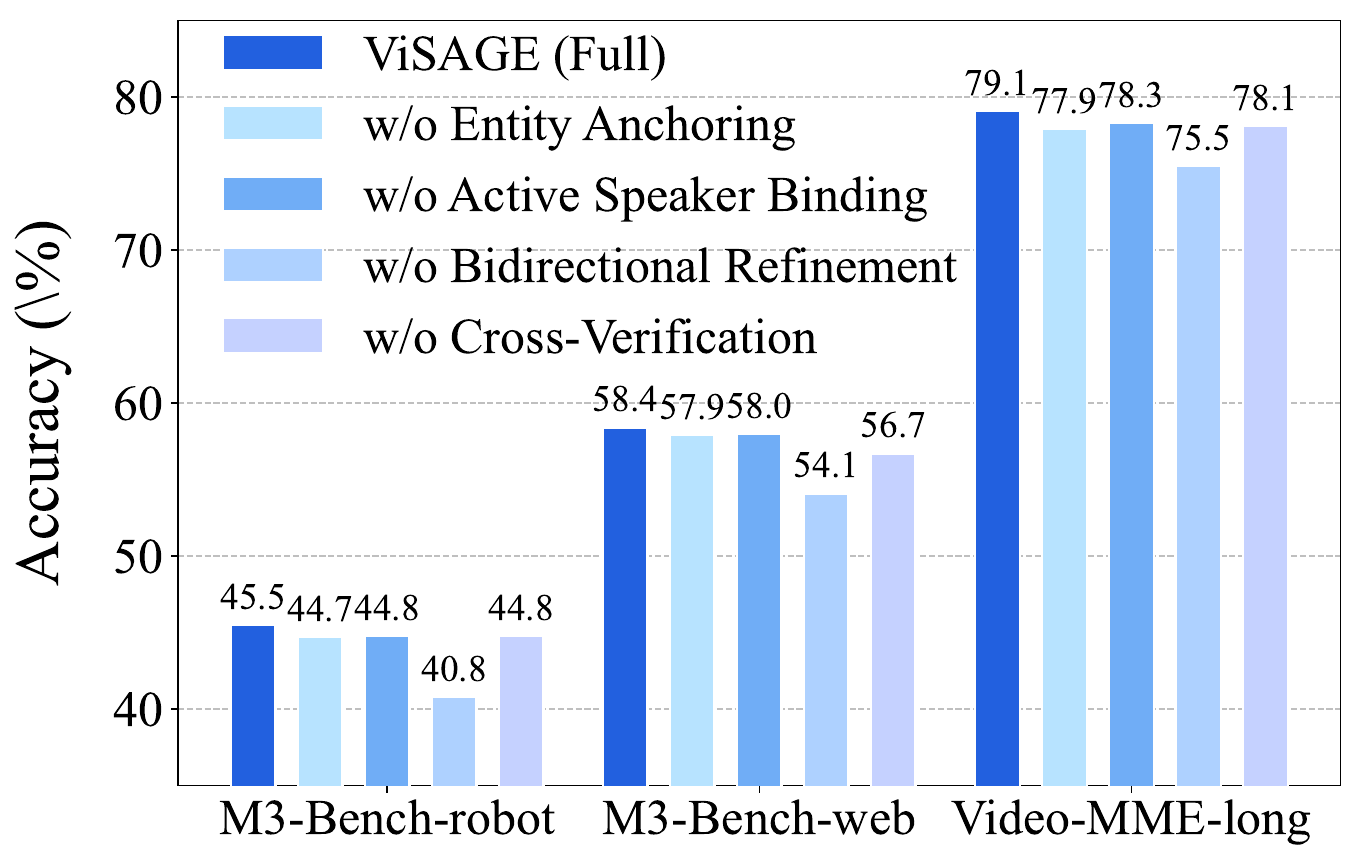}
    \caption{Ablation study on key components. The chart illustrates the impact of removing each module across three benchmarks (M3-Bench-robot, M3-Bench-web, and Video-MME-long).}
    \Description{Ablation study on key components. The chart illustrates the impact of removing each module across three benchmarks (M3-Bench-robot, M3-Bench-web, and Video-MME-long).}
    \label{fig:ablation}
\end{figure}

\subsection{Performance Analysis}
Table~\ref{tab:main} presents the quantitative evaluation. \ourmethod consistently demonstrates superior efficacy across all benchmarks.

Crucially, regarding the Cross-Modal metric, \ourmethod achieves a decisive breakthrough. This metric directly validates the resolution of the Spatio-Temporal Detail Loss, measuring the system's ability to bind fragmented visual and auditory cues into coherent entities. On this front, \ourmethod significantly outperforms the agentic baseline M3-Agent (improving +15.9\% on robot and +27.1\% on web), proving that our Entity Anchoring mechanism successfully counters signal degradation where standard agents fail.

Furthermore, \ourmethod grounds the raw capabilities of the Gemini-3-Pro backbone. While the backbone exhibits strong parametric knowledge, it remains vulnerable to the sampling-induced information loss, which scores only 38.0\% on robot CM. By shifting from passive perception to entity-anchored reasoning, \ourmethod bridges this gap, boosting the score to 47.1\%. This robust grounding capability extends to open-world long-horizon scenarios, where \ourmethod achieves a new state-of-the-art (SOTA) accuracy of 79.1\% on Video-MME-long, surpassing both the ungrounded backbone (74.2\%) and the fragmentation-prone M3-Agent (61.8\%). Because Video-MME is widely recognized as a comprehensive and highly generalized benchmark in the long video understanding domain, achieving SOTA here validates our method's strong adaptability to diverse, open-world video content. Admittedly, constrained by current off-the-shelf perception tools (i.e., face and voiceprint recognition), our implementation is presently restricted to human-centric video understanding. Nevertheless, tracking and reasoning about human dynamics is arguably the most prevalent and critical application scenario in this field, laying the essential groundwork for real-world deployments such as embodied AI and smart assistants.

\begin{table}[t]
    \centering
    \caption{Impact of backbone scaling. We report the average accuracy on M3-Bench-robot~\citep{long2025seeing}.}
    \Description{Impact of backbone scaling. We report the average accuracy on M3-Bench-robot~\citep{long2025seeing}.}
    % 移除了 \resizebox，直接使用自然排版，默认继承正文的 9pt 字号
    \begin{tabular}{lc}
        \toprule
        \textbf{Backbone Model} & \textbf{Accuracy (\%)} \\
        \midrule
        \ourmethod w/ Qwen3-VL-30b & 42.1 \\
        \ourmethod w/ GPT-5 & 44.5 \\
        \ourmethod w/ Gemini-3-Pro & 45.5 \\
        \bottomrule
    \end{tabular}

    \label{tab:backbone_ablation}
    % 只要没有像之前那样造成文字重叠，保留它稍微收缩一点表格下方的间距是允许的。 
\end{table}

\subsection{Ablation Study}
Fig.~\ref{fig:ablation} and Table~\ref{tab:backbone_ablation} summarize our component analysis.
Structurally, Bidirectional Memory Refinement proves critical (Fig.~\ref{fig:ablation}); its removal triggers the sharpest drop ($-4.7\%$ on robot), confirming that Backward Rectification is essential for unifying the fragmented narratives caused by the Segmentation Dilemma.
Other modules also contribute distinctively to the system's robustness: Entity Anchoring (Part A) provides consistent gains by securing identity stability against the spatio-temporal detail loss. Multi-Agent Cross-Verification (Part C) further boosts precision by filtering semantic noise and enforcing epistemic safety. 
Crucially, Table~\ref{tab:backbone_ablation} highlights the universality of our framework. \ourmethod maintains robust performance across distinct architectures ranging from Qwen3-VL-30b to GPT-5~\citep{openai2025gpt5}. This validates that our Dual-Track design is a model-agnostic framework: it successfully decouples memory structure from specific reasoning backends, enhancing any foundation model with long-horizon reasoning capabilities.

\begin{table}[htb]
    \centering
    \small % 保持 8pt，字号合规
    \renewcommand{\arraystretch}{1.3} 
    
    % 【1. 标题置顶】
    \caption{Qualitative Case Study. \ourmethod successfully resolves the ambiguous pronoun ``I'' via retroactive refinement, whereas the baseline fails due to surface-level matching.}
    \label{tab:case_study}
    
    % 【2. 去掉 resizebox，使用 0.96\linewidth 完美适配单栏宽度】
    \begin{tabular}{p{0.96\linewidth}}
    \toprule
    % 【3. 补充一个表头以符合无障碍语义规范】
    \textbf{Evaluation Aspect \& Reasoning Chain} \\ 
    \midrule % <--- 全场唯一合法的 midrule
    
    \textbf{User Query:} ``What is Emma's major?'' \newline
    \textbf{Ground Truth:} Programming. \\
    \addlinespace[1ex] % 用垂直留白代替 midrule
    
    % 【4. 优化灰底色块：将 baseline 内容合并在一个单元格内，用 \newline 换行，防止灰底断层】
    \rowcolor{gray!10} 
    \textbf{Gemini Response (Baseline):} \newline
    ``Emma's major is \textbf{Linear Algebra} \textcolor{red}{\ding{55}}.'' \newline
    \textit{\textbf{Retrieved Memory:}} A young woman says: ``\textit{I know Emma is good at Linear Algebra.}'' \newline
    \textit{\textbf{Error Analysis:}} The baseline relies on surface-level name matching, prioritizing the explicit mention of ``Emma'' by a third party while ignoring the pronoun ``I'' spoken by the target character (who was initially anonymous). \\
    \addlinespace[1.5ex] % 用垂直留白代替 midrule
    
    \textbf{\ourmethod (Ours) Reasoning Process:} \newline
    \textbf{1. Bidirectional Refinement (Resolving ``I''):} \newline
    \hspace*{1em}\textbullet~\textit{Initial State:} At [00:19], \texttt{<p\_7>} says ``\textbf{I} study Programming.'' (Speaker unknown). \newline
    \hspace*{1em}\textbullet~\textit{Trigger \& Update:} At [05:30], identity revealed (\texttt{<p\_7>} $\rightarrow$ Emma). System \textbf{retroactively propagates} this identity, modifying the historical log at [00:19] to: ``\textbf{p\_7=Emma} says `I study Programming.''' \newline
    \textbf{2. Cross-Verification:} \newline
    \hspace*{1em}\textbullet~\textbf{Character Agent (Emma):} ``My major is \textbf{Programming}. I recall stating this at the start of the conversation.'' \newline
    \hspace*{1em}\textbullet~\textbf{Scene Agent (Objective):} Retrieves two relevant logs: \newline
    \hspace*{2em}1. [00:19] \textbf{Emma} says: ``I study Programming.'' \newline
    \hspace*{2em}2. [02:40] \textbf{Lily} says: ``...I know Emma is good at Linear Algebra.'' \newline
    \hspace*{1em}\textbullet~\textbf{Judge Verdict:} ``Evidence 1 is a \textbf{direct first-person statement}, whereas Evidence 2 is \textbf{hearsay}. Direct claim takes precedence. \textbf{Final Answer: Programming \textcolor{teal}{\ding{51}}}'' \\ 
    \bottomrule
    \end{tabular}
\end{table}

\begin{figure}[!t] 
    \centering 
    \includegraphics[width=0.9\linewidth]{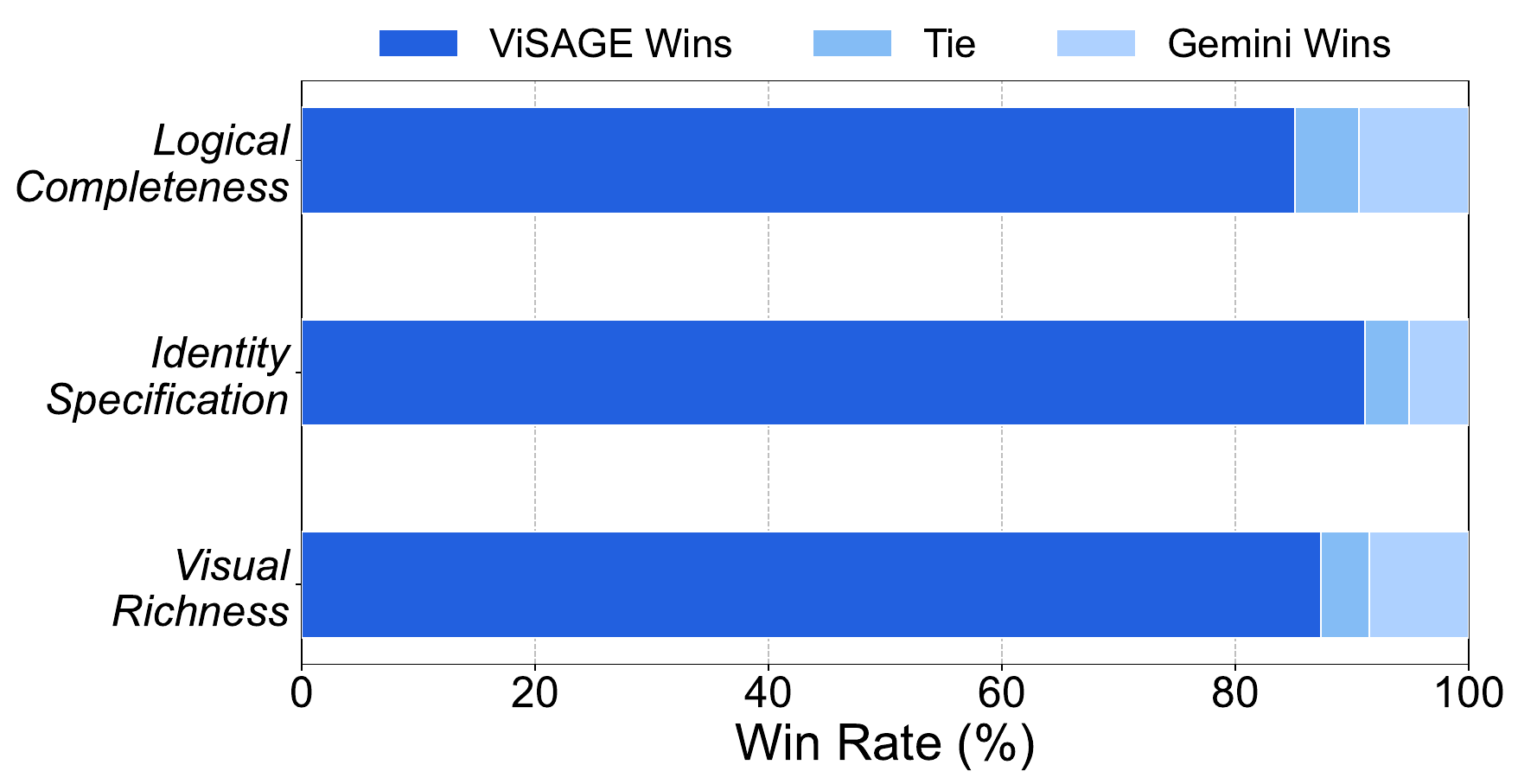}
    \caption{Quality Assessment on Correct Samples. Even when the final answer is correct, \ourmethod generates responses with significantly higher visual richness and logical completeness compared to baselines.}
    \Description{Quality Assessment on Correct Samples. Even when the final answer is correct, \ourmethod generates responses with significantly higher visual richness and logical completeness compared to baselines.}
    \label{fig:win_ana}
\end{figure}

\subsection{Qualitative and Safety Analysis}

\paragraph{Case Study: Resolving Pronominal Ambiguity.} As detailed in Table~\ref{tab:case_study}, the baseline fails on pronominal resolution, succumbing to misleading hearsay. In contrast, \ourmethod successfully disentangles this ambiguity. By retroactively grounding the first-person pronoun (``I'') to the target entity, our system enables the Judge Agent to prioritize direct testimony over third-person claims, effectively preventing hallucination.

\paragraph{Beyond Accuracy: Response Granularity.} 
Fig.~\ref{fig:win_ana} reveals that correct labels do not imply full comprehension. 
Employing GPT-5 as a judge, we evaluated three dimensions: 
(1) \textbf{Visual Richness} (density of scene details), where \ourmethod achieves a win rate of 87.3\%; 
(2) \textbf{Identity Specification} (pinpointing who performs the action), achieving our highest margin of 91.1\%; 
(3) \textbf{Logical Completeness} (providing transparent reasoning steps), with an 85.1\% lead.
These gaps confirm that while baselines rely on shallow pattern matching (correct label, sparse evidence), \ourmethod constructs grounded narratives with precise entity binding.
\begin{figure}[t] 
    \centering 
    \includegraphics[width=0.9\linewidth]{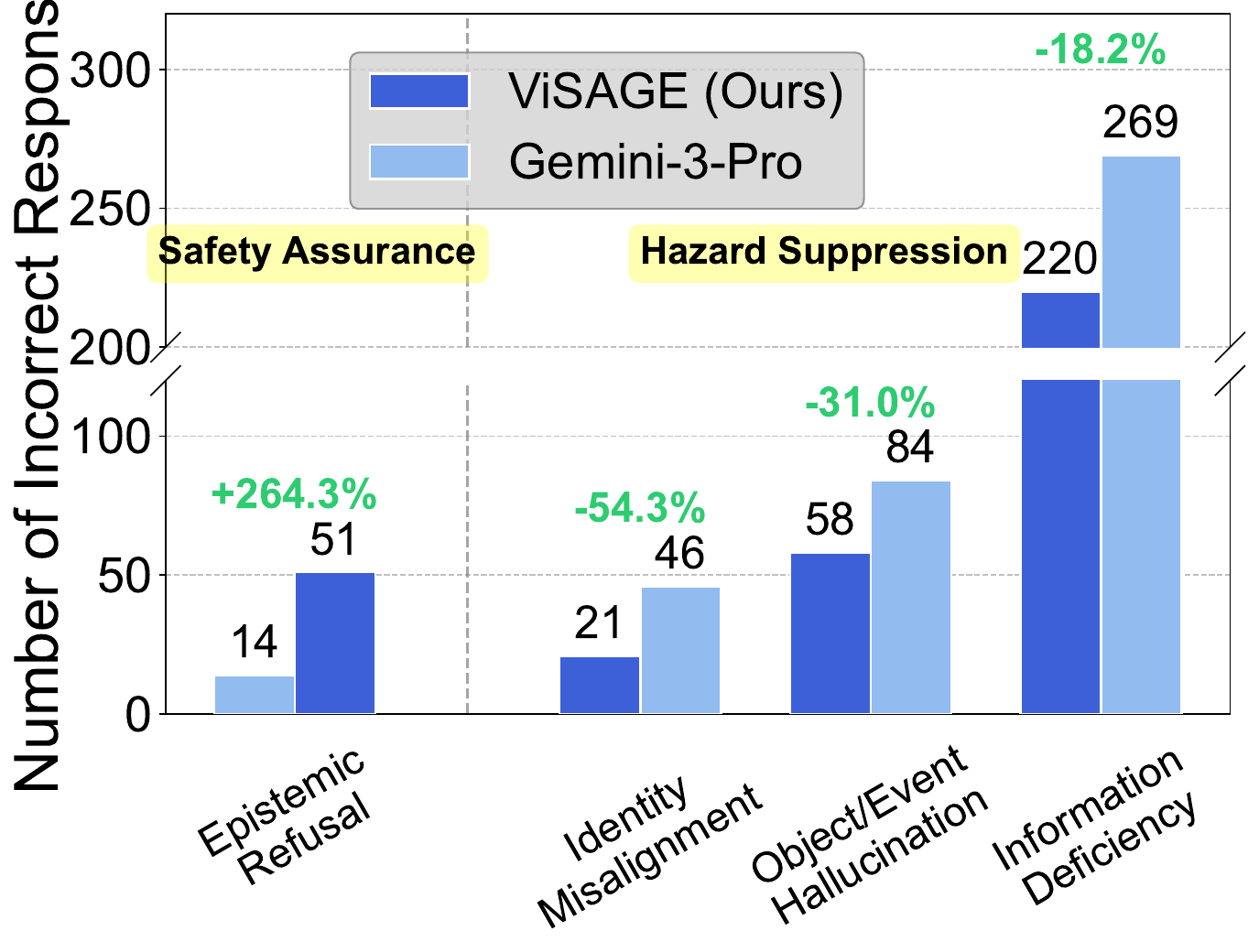} 
    \caption{Safety Analysis on M3-Bench-robot. \ourmethod significantly reduces hazardous hallucinations and increases safe epistemic refusals.}
    \Description{Safety Analysis on M3-Bench-robot. \ourmethod significantly reduces hazardous hallucinations and increases safe epistemic refusals.}
    \label{fig:error_analysis}
\end{figure}
\paragraph{Failure Mode Analysis: The Safety of ``Knowing What You Don't Know.''} 
Finally, we analyze failure boundaries by categorizing errors into Hazardous Hallucinations and Safe Refusals. 
Fig.~\ref{fig:error_analysis} reveals a critical shift in \ourmethod, prioritizing safety in high-stakes scenarios:

\noindent (1) \textbf{Mitigating Hazardous Failures.} The most dangerous errors in embodied agents are Identity Misalignment (wrongly attributing actions to specific individuals) and Object/Event Hallucination (fabricating non-existent entities). 
\ourmethod significantly suppresses these risks, reducing Identity Misalignment by 54.3\% and Hallucinations by 31.0\%, thereby preventing \textbf{execution} based on false premises. 

\noindent (2) \textbf{Embracing Safe Refusals.} Instead of hallucinating, \ourmethod exhibits a dramatic 264.3\% increase in Epistemic Refusal: the explicit acknowledgment of information deficit. 
In robotics, this represents a \emph{graceful degradation}: preferring an honest ``I don't know'' over a confident error is vital for trustworthiness and safety.

\subsection{Cost and Latency Analysis}
\label{sec:cost}

\begin{figure}[t]
  \centering
  \includegraphics[width=\linewidth]{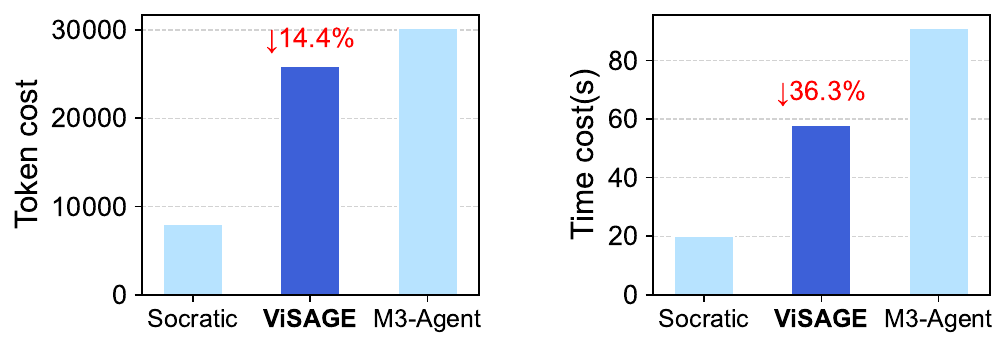}
  \caption{Cost and latency comparison among the Socratic method, \ourmethod, and M3-Agent for processing a 30-second video segment. \ourmethod achieves superior accuracy with less overhead compared to the previous SOTA agentic framework.}
  \Description{Cost and latency comparison among the Socratic method, \ourmethod, and M3-Agent for processing a 30-second video segment. \ourmethod achieves superior accuracy with less overhead compared to the previous SOTA agentic framework.}
  \label{fig:cost_latency}
\end{figure}

In this section, we analyze the computational and temporal overhead of our proposed framework. We conduct the evaluation on the M3-Bench-robot dataset~\citep{long2025seeing}, utilizing Gemini-3-Pro~\citep{gemini3_2025} as the backbone model. We compare \ourmethod against the standard Socratic method~\citep{zeng2022socratic} and the prior state-of-the-art agentic memory framework, M3-Agent~\citep{long2025seeing}. Since all three approaches process long-form videos by segmenting them into localized chunks, we standardize our measurement based on the average cost to process a 30-second video clip. The comparative results are illustrated in Figure~\ref{fig:cost_latency}.

The \ourmethod processing pipeline consists of two primary stages. The first stage employs lightweight perception tools for object detection and TalkNet~\citep{tao2021someone} for active speaker detection, to perform entity-anchored preprocessing. Because these specialized tools are significantly smaller than MLLMs, their inference latency is almost negligible, typically taking only around 10 milliseconds per frame. Consequently, the primary computational bottleneck---and the focus of our comparison---lies in the second stage: the LLM execution phase.

In the second stage, the backbone MLLM processes the visual frames alongside the extracted entity cues to construct the memory bank. As shown in Figure~\ref{fig:cost_latency}, the pure Socratic method provides a highly efficient baseline ($\approx$8,000 tokens and $\approx$20s) because it relies on a single, straightforward visual perception pass without maintaining an evolving memory structure. In contrast, agentic frameworks inherently necessitate higher token consumption to manage and retrieve memory. However, thanks to our robust sequence-level entity anchoring during preprocessing, \ourmethod can aggressively downsample the video input to 0.5 frames per second while preserving crucial identity dynamics. This allows \ourmethod to achieve superior reasoning performance without a catastrophic explosion in overhead. 

Specifically, \ourmethod consumes approximately 26,000 tokens and takes $\approx$60s per 30-second clip. This is notably more efficient than M3-Agent, which requires $\approx$30,000 tokens and $\approx$90s. The elevated latency in M3-Agent is largely driven by the heavy communication overhead of transmitting raw video segments to the multimodal API, compounded by its complex, multi-round graph retrieval and semantic matching processes. By resolving identities directly upon evidence discovery rather than relying on massive video transmission and exhaustive multi-hop retrieval, \ourmethod successfully balances high-fidelity long-video understanding with practical computational efficiency.

\section{Conclusion and Limitations}

In this paper, we argue that long-form video understanding requires a paradigm shift from strictly sequential processing to a self-correcting memory architecture. By enabling agents to retroactively align historical memory with evolving identity evidence, \ourmethod demonstrates that post-hoc rectification is a structural necessity for maintaining narrative coherence. Beyond accuracy, our framework establishes a new standard for epistemic safety in embodied AI: rather than prioritizing plausible fabrication, it enforces rigorous evidence verification. This transition from hazardous hallucinations to verified refusals is critical for deploying trustworthy agents in real-world environments, ensuring that future systems ``know what they don't know.''

Although \ourmethod shows notable improvements, we note a primary limitation in the current system. Our Entity-Anchored Preprocessing is explicitly tailored for human-centric interactions, utilizing \texttt{Face\_ID} and \texttt{Speaker\_ID} bindings to resolve complex social dynamics. Consequently, the system presently treats non-human elements, such as plot-critical objects or animals, as background context rather than actively tracked profiles. To address this, our future work aims to generalize the scope of entity registration by incorporating open-vocabulary object tracking and re-identification. This will evolve \ourmethod from a character-centric assistant into a broadly entity-aware agent.

\bibliographystyle{ACM-Reference-Format}
\bibliography{main}

\end{document}